\documentclass[fleqn,10pt]{wlscirep}
\usepackage[utf8]{inputenc}
\usepackage[T1]{fontenc}
\title{A novel spatial-frequency domain network for zero-shot incremental learning}

\author[1]{Jie Ren}
\author[1]{Yang Zhao}
\author[2,*]{Weichuan Zhang}
\author[3]{Changming Sun}
\affil[1]{Xi'an Polytechnic University, Xi'an, Shaanxi Province, China}
\affil[2]{Shaanxi University of Science and Technology, Xi 'an, Shaanxi Province, China}
\affil[3]{CSIRO Data61, PO Box 76, Epping, NSW 1710, Australia.}

\affil[*]{Corresponding author: Weichuan Zhang, zwc2003@163.com}

\begin{abstract}
	Zero-shot incremental learning aims to enable the model to generalize to new classes without forgetting previously learned classes. However, the semantic gap between old and new sample classes can lead to catastrophic forgetting. Additionally, existing algorithms lack capturing significant information from each sample image domain, impairing models' classification performance. Therefore, this paper proposes a novel Spatial-Frequency Domain Network (SFDNet) which contains a Spatial-Frequency Feature Extraction (SFFE) module and Attention Feature Alignment (AFA) module to improve the Zero-Shot Translation for Class Incremental algorithm. Firstly, SFFE module is designed which contains a dual attention mechanism for obtaining salient spatial-frequency feature information. Secondly, a novel feature fusion module is conducted for obtaining fused spatial-frequency domain features. Thirdly, the Nearest Class Mean classifier is utilized to select the most suitable category. Finally, iteration between tasks is performed using the Zero-Shot Translation model. The proposed SFDNet has the ability to effectively extract spatial-frequency feature representation from input images, improve the accuracy of image classification, and fundamentally alleviate catastrophic forgetting. Extensive experiments on the CUB 200-2011 and CIFAR100 datasets demonstrate that our proposed algorithm outperforms state-of-the-art incremental learning algorithms.
\end{abstract}
\begin{document}
	
	\flushbottom
	\maketitle
	%
	%
	\thispagestyle{empty}
	
	\noindent keywords: Catastrophic forgetting, incremental learning, zero-shot learning, convolutional neural networks, discrete cosine transform, attention mechanism
	
	\section*{Introduction}
	
\begin{figure}[ht]
		\begin{center}
			\includegraphics[width=1.0\linewidth]{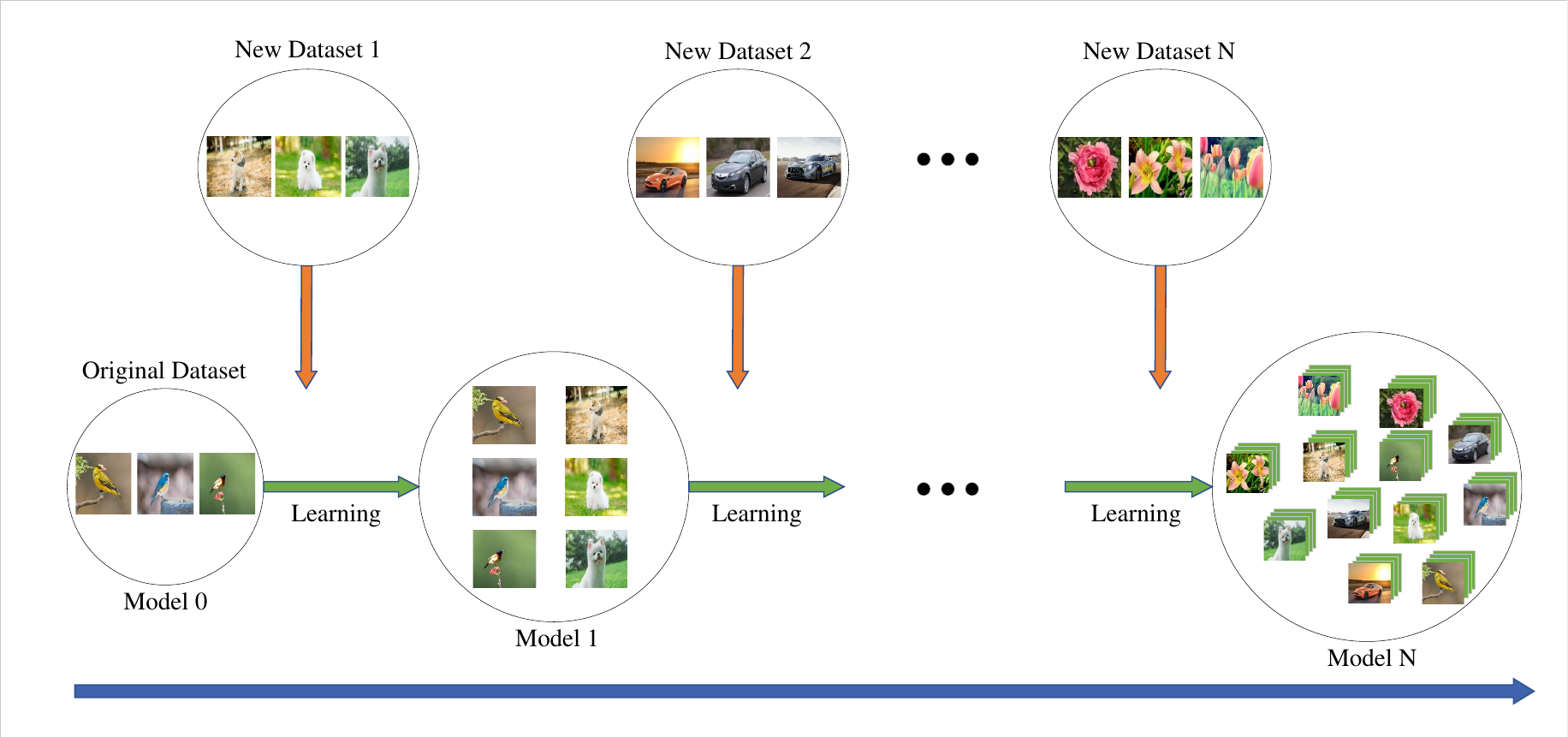}
		\end{center}
		\caption{\label{Figure 1}Incremental Learning}
	\end{figure}

	In recent years, machine learning \cite{yang2018new,zhang2024re}, using computers to simulate or realize human learning activities~\cite{zhang2021ndpnet, jing2022image, zhang2023image,jing2023ecfrnet}, has made remarkable achievements. However, unlike human beings who have the ability to continuously learn new knowledge without forgetting old knowledge, after a machine learning model is trained with a new dataset, its feature distribution will be biased towards the feature distribution of the new dataset, leading to a significant drop in performance on the old dataset. This phenomenon is called catastrophic forgetting \cite{robins1995catastrophic} and has become one of the hottest research topics in the field of artificial intelligence.
	The main reason for catastrophic forgetting is that the data distribution is assumed to be fixed or smooth, and the training samples are under the assumption to be independently and identically distributed before training traditional models. Therefore, the model can learn knowledge from the same data during repeated tasks. However, when the input dataset becomes a continuous stream, its distribution is non-smooth and dependent. The new knowledge will interfere with the old knowledge the model learned previously, leading to a rapid degradation of the model's performance on the old dataset.
	To alleviate catastrophic forgetting, researchers focus on improving the plasticity and stability of the model. The intuitive way is to use all data to retrain the model to adapt to changes in data distribution. Although this approach can completely solve the catastrophic forgetting problem, its inefficiency restricts the model learning in real time.
	Considering limited computation and storage, Incremental learning \cite{1989Catastrophic}, also known as continuous learning or Lifelong Learning, which enables machines to continuously acquire, adjust, and transfer knowledge like humans, is proposed to balance plasticity and stability requirements. With incremental learning (shown in Figure \ref{Figure 1}), the network continuously processes the continuous flow in the real world, retaining and even integrating and optimizing the old knowledge while absorbing new knowledge.
	
	Existing incremental learning methods can be divided into three categories: (1) Regularization-based incremental learning, such as LWF \cite{li2017learning}, EWC \cite{kirkpatrick2017overcoming}, and MAS \cite{aljundi2018memory}. These methods protect old knowledge from being overwritten by new knowledge by adding constraints to the loss function of the new task and usually do not require old data; (2) Replay-based incremental learning, such as iCaRL \cite{2016iCaRL}, DGR \cite{shin2017continual}, and GEM \cite{lopez2017gradient}, follows a "learning from the past" strategy. Because part of the representative old data is selected and used by the model to review old knowledge during training for a new task, it is necessary to consider which part of the data in the old task is representative and how to retrain the model with the new and selected old data together; (3) Parameter-isolation based incremental learning, such as PNN \cite{rusu2016progressive}, DAN \cite{rosenfeld2018incremental}, and HAT \cite{serra2018overcoming}, solves the catastrophic forgetting problem by assigning new parameters to new tasks. Due to the introduction of a large number of parameters and computational complexity, these methods are usually used in simpler incremental learning tasks. Since regularization-based incremental learning methods are closer to the real goal of incremental learning, they constitute the main research direction of incremental learning.
	
	To avoid catastrophic forgetting caused by the added new data in the new task, researchers focus on narrowing the distance between the new task and the old task. Firstly, the model is retrained without using old data, thereby improving image classification accuracy while maintaining retraining efficiency. Li et al. \cite{li2017learning} proposed the Learning without Forgetting (LWF) model, which utilizes only new data and knowledge distillation to train the model while preserving capabilities similar to the old model. It is the milestone of incremental learning. To better improve the robustness and generalization of the model, Zhang et al. \cite{zhang2020class} proposed the Deep Model Consolidation (DMC) model, which first trains a separate model for the new class and then combines two models trained by both new and old tasks through a double distillation training objective. The Less-forgetting Learning (LFL) model proposed by Jung et al. \cite{jung2016less} uses stochastic gradient descent, a more flexible fine-tune method, to better adapt to new tasks and effectively relieve catastrophic forgetting caused by old samples. However, the above methods will lead to an increase in computation and storage costs \cite{de2019continual}.
	
	On the other hand, to better reduce the distance between old and new tasks, researchers utilize the parameter distribution of old tasks to train the network model. Inspired by the LWF model, James et al. \cite{kirkpatrick2017overcoming} proposed a parameter-related regularization loss in the Elastic Weight Consolidation (EWC) model, which guides the parameters of the new model trained on the new task to be as close as possible to the parameters of the old model based on the importance of parameters. However, EWC is usually used in static environments and is not suitable for dynamic environments or rapidly evolving tasks \cite{de2019continual}. Lee et al. \cite{lee2017overcoming} proposed the Incremental Moment Matching (IMM) model, which incrementally matches two moments of the posterior distribution of the neural network trained on two adjacent tasks. As IMM uses a large amount of old task data in the incremental matching, its performance may decline when the amount of data is limited \cite{de2019continual}. Additionally, the Memory Aware Synapses (MAS) model proposed by Rahaf et al. \cite{aljundi2018memory} first estimates importance weights for all the network parameters relying on the sensitivity of the predicted output function. Then, the parameters with high importance are selected to remain unchanged in the subsequent learning process. Because of the accumulation of importance measures for each parameter of the network, the MAS model requires additional storage, leading to an increase in memory and storage costs \cite{de2019continual}.
	
\begin{figure}[ht]
		\begin{center}
			\includegraphics[width=1.0\linewidth]{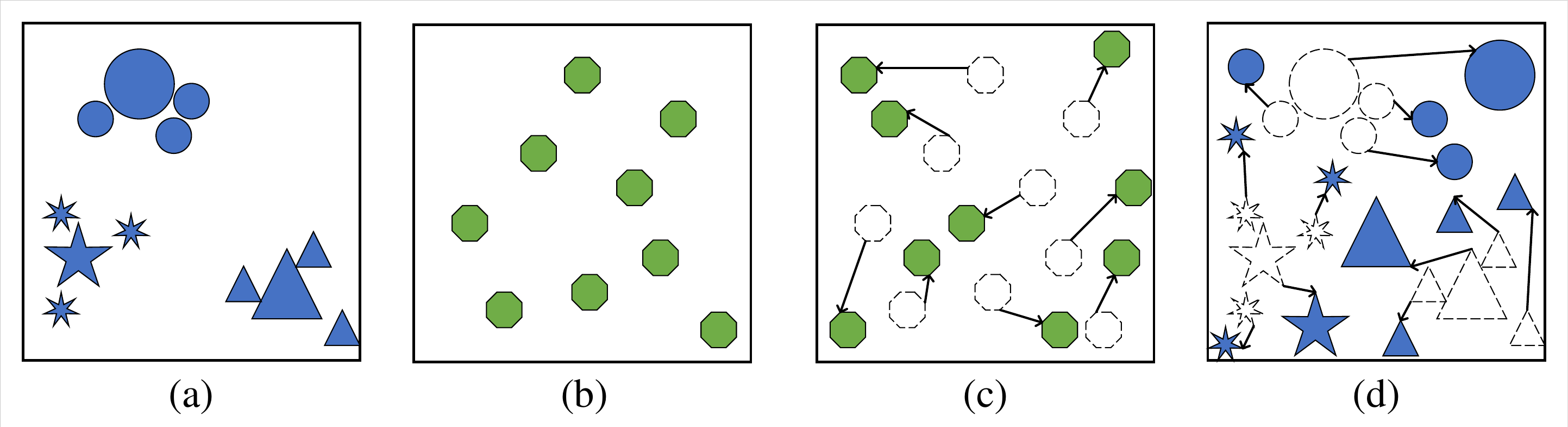}
		\end{center}
		\caption{\label{Figure 2}Schematic diagram of the process of data and prototypes shift during training with new task.}
	\end{figure}

	With further research, scholars found that the distance between new and old tasks is mainly caused by the semantic gap \cite{2011Semantic} between embedding spaces. Figure \ref{Figure 2} shows a schematic diagram of how the semantic gap is generated, where: (a) represents the data and prototypes of three classes in task 1 after training task 1, where small graphs represent data and large graphs represent prototypes, which is the class embedding mean; (b) represents the data of task 2 that needs to be added in training task 2; (c) represents the data drift of task 2 during training task 2; (d) represents the drift of data and prototypes of three classes in task 1 after training task 2. It illustrates that the data and prototypes of each class will deviate after training for Task 2, causing a decline in the performance of the model for Task 1. Moreover, as the number of tasks increases, the data and prototype of each class in the embedding space will experience more drift, thus leading to an increasing distance between old and new tasks. Therefore, the performance of the latest trained model on old tasks drops significantly, generating catastrophic forgetting.
	
	In order to solve the semantic gap between two embedding spaces and reduce the distance between old and new samples, Yu et al. \cite{yu2020semantic} proposed Semantic Drift Compensation (SDC) with a semantic compensation method. This model introduces a triple loss function to estimate the center of each category without storing images. Classification prediction is then performed through the nearest neighbor class mean classifier (NCM) \cite{kang2019decoupling}. Although the SDC model can compensate for semantic gaps without requiring any old sample data, it ignores the relationship between classes from different tasks, directly affecting the performance of semantic gap compensation \cite{wei2021incremental}. Inspired by the SDC model, to more effectively address the semantic gap problem, Wei et al. \cite{wei2021incremental} combined zero-shot learning \cite{lampert2009learning} with incremental learning to generate a Zero-Shot Translation Class-Incremental (ZSTCI) model. Zero-shot learning refers to the process in which the model can use the knowledge of visible categories to identify unseen categories without training samples. ZSTCI can construct a zero-shot translation between two adjacent tasks, estimating and compensating for the semantic gap of two adjacent embedding spaces. It then learns a unified representation in a common embedding space, enabling the model to accurately capture the relationship between the previous class and the current class. Although the above models can significantly improve the semantic gap problem, they are unable to identify the basic cause of the semantic gap \cite{dhar2019learning}. Based on this research, Wei et al. \cite{wei2021incremental} proposed using a generative regression strategy to transform zero-shot incremental learning into traditional zero-shot learning. Simultaneously, they employed a distillation strategy to extract information from the previous model into the current task, thereby mitigating catastrophic forgetting. Additionally, building upon zero-shot incremental learning, Sun et al. \cite{sun2023class} introduced Class-Incremental Generalized Zero-Shot Learning (CIG-ZSL) as an extension of generalized zero-shot learning. They introduced the Dual Path Learner (DPL) and demonstrated the feasibility of addressing CIG-ZSL tasks on coarse-grained and fine-grained datasets.
	
	In 2019, an attention distillation loss designed in the Learning without Memorizing (LWM) \cite{dhar2019learning} model enables the network to focus on local information, thereby enhancing useful information and suppressing useless information \cite{mnih2014recurrent}. After extensive experimental verification, this work proves that one of the fundamental causes of semantic gaps is attention bias. Inspired by LWM, Ding \cite{ding2022incremental} added a channel attention mechanism to the network to better resist forgetting old categories.
	Due to its resource-saving and flexible advantages, research on zero-shot incremental learning has become a major trend in mitigating catastrophic forgetting issues. However, the image classification performance of zero-shot incremental learning is still less than satisfactory. A crucial reason contributing to this problem is the attention shift during the incremental learning process, leading to a decline in the model's performance over time when dealing with new classes. A crucial factor contributing to this is the attention shift during the incremental learning process, resulting in a decline in classification performance on old tasks. However, a single attention mechanism not only fails to capture global contextual information effectively but also exhibits significant performance differences across different sample images. These make it challenging to express the rich features of sample images, thereby suppressing sample diversity.
	Furthermore, although the use of convolutional neural networks for extracting features in the spatial domain has been extensively studied, to extract a more comprehensive and richer feature representation of images, some scholars propose the supplementation of spatial information with frequency domain information, which has demonstrated notable performance enhancements in various image-related tasks~\cite{zhang2014corner, jing2022recent, zhang2019corner, zhang2020corner, zhang2017noise, zhang2019discrete,shui2013corner,liao2022asrsnet, zhang2015contour,shui2012noise}.
	
	After transforming images from the spatial domain to the frequency domain through methods such as Fourier transform and discrete cosine transform, features of different scales and orientations become more distinguishable. These features can be utilized for the recognition and classification of detailed information. The frequency domain attention mechanism \cite{qin2021fcanet}, which enables the model to give more attention to crucial frequency components in the image, can enhance the robustness and recognizability of the images. However, existing zero-shot incremental learning which relies solely on extracting image features from only spatial domain for image classification, fails to capture diverse features across different domains, resulting in a reduction of network robustness, generalization capabilities, and a decrease in image classification accuracy.

\begin{figure}[ht]
		\begin{center}
			\includegraphics[width=1.0\linewidth]{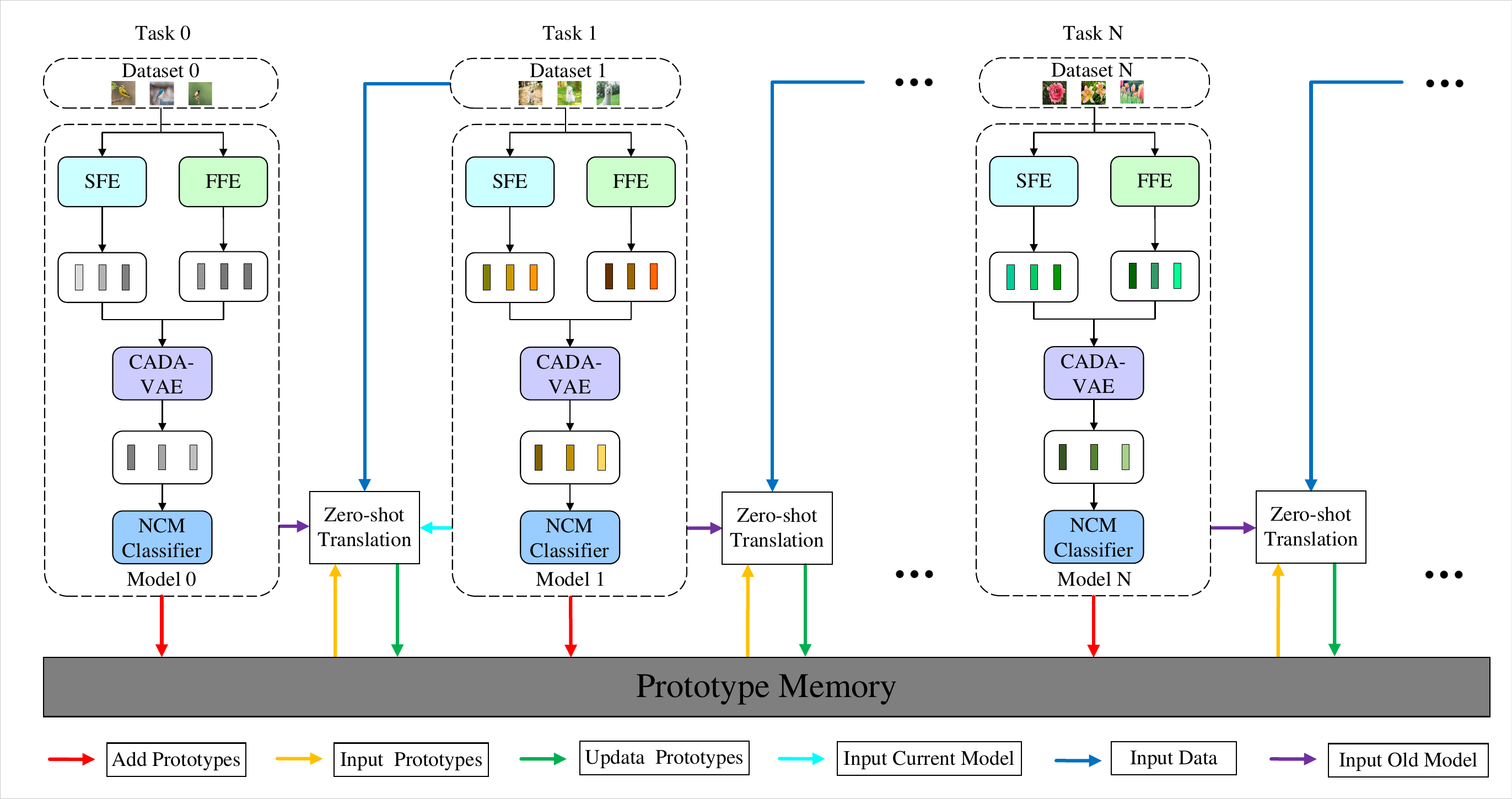}
		\end{center}
		\caption{\label{Figure 3}The architecture of SFDNet.}
	\end{figure}

	To address the aforementioned problems, this paper proposes a novel Spatial-Frequency Domain Network (SFDNet) for zero-shot incremental learning, which includes a Spatial-Frequency Feature Extraction (SFFE) module and Attention Feature Alignment (AFA) module to improve the Zero-Shot Translation for Class Incremental algorithm, as illustrated in Figure \ref{Figure 3}. Since Yu et al. \cite{yu2020semantic} demonstrated that the embedding network is less affected by catastrophic forgetting, this paper mainly focuses on improving the feature extraction module, enabling the network to obtain more useful feature information and thereby enhancing the classification performance of the new model on old tasks. Furthermore, considering that in real-life situations, humans tend to focus on the crucial parts and main features of a new object when observing it, approaching the issue comprehensively from various perspectives. In the Spatial-Frequency Feature Extraction (SFFE) module, the spatial feature extraction (SFE) module and the frequency feature extraction (FFE) module are designed to extract the spatial and frequency features of the images, respectively. The attention feature alignment (AFA) module based on the spatial-frequency domain is proposed to obtain more prominent spatial and frequency features on the SFE and FFE modules. The obtained spatial and frequency features are first aligned through a cross-distribution alignment mechanism and then fused to generate the image features. Then, these image features are fed into the Nearest Class Mean (NCM) classifier for classification. Finally, iteration between tasks is performed using the Zero-Shot Translation model. This SFDNet enables machines to "think" more like humans, effectively improving image classification accuracy and alleviating catastrophic forgetting in the network. Additionally, during the continuous learning process, this paper employs a zero-shot translation module to compensate for the semantic gaps between new and old feature embeddings. A cross-alignment loss is designed to mitigate the impact of catastrophic forgetting.
	
	The main contributions of this paper are as follows:
	
	1) In order to obtain richer image features, this paper proposes a novel Spatial-Frequency Domain Network (SFDNet) which contains a Spatial-Frequency Feature Extraction (SFFE) module and Attention Feature Alignment (AFA) module to improve the Zero-Shot Translation for Class Incremental algorithm. The SFFE module mainly includes the spatial feature extraction (SFE) module and the frequency feature extraction (FFE) module, which are used to obtain the spatial and frequency domain features of the image respectively. The AFA module combined spatial and frequency domain attention mechanisms to enhance the network by focusing on areas of interest, improving the effectiveness of image feature extraction.
	
	2) Considering the distinctions between spatial and frequency domain features, this paper pioneers the integration of a frequency domain feature extraction network into zero-shot incremental learning. This enables the network to obtain more comprehensive image feature information, effectively enhancing image classification accuracy, and alleviating the catastrophic forgetting in the network.
	
	\section*{Method}
	
	To compensate for the semantic gap between two tasks more effectively, a novel zero-shot incremental learning method based on spatial-frequency domain attention feature alignment is proposed, which includes five modules: spatial domain feature extraction network, frequency domain feature extraction network, spatial-frequency domain attention feature alignment network, spatial-frequency domain feature fusion network, and zero-shot translation. We extract significant image features from both spatial and frequency domains, guiding the compensation of semantic gaps in zero-shot translation.
	
	\subsection*{Spatial-Domain Feature Extraction Module: SFE}
	
\begin{figure}[ht]
		\begin{center}
			\includegraphics[width=1.0\linewidth]{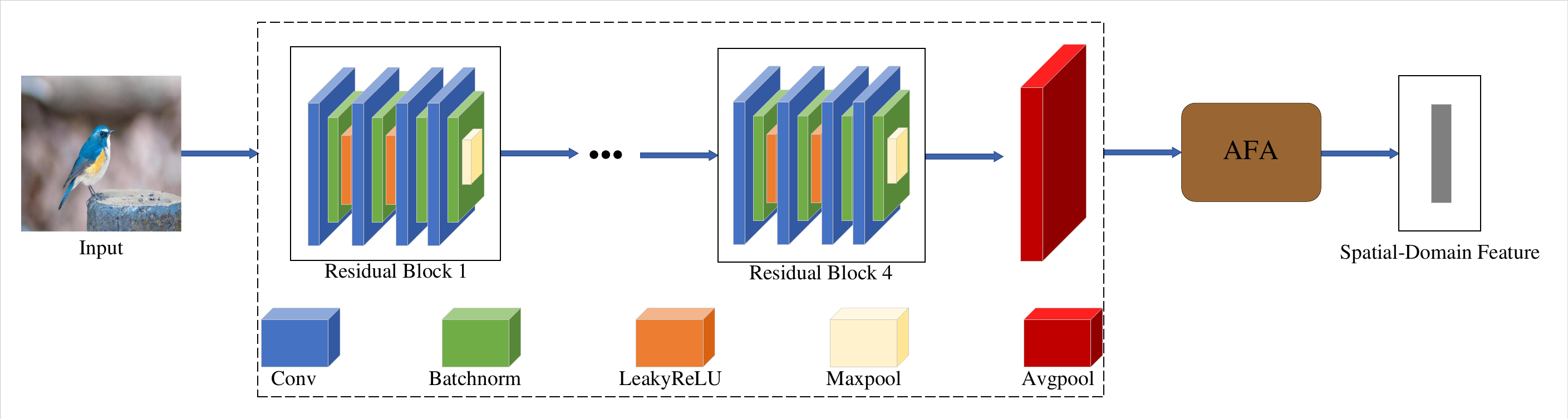}
		\end{center}
		\caption{\label{Figure 4}SFE module.}
	\end{figure}

\begin{figure}[ht]
		\begin{center}
			\includegraphics[width=1.0\linewidth]{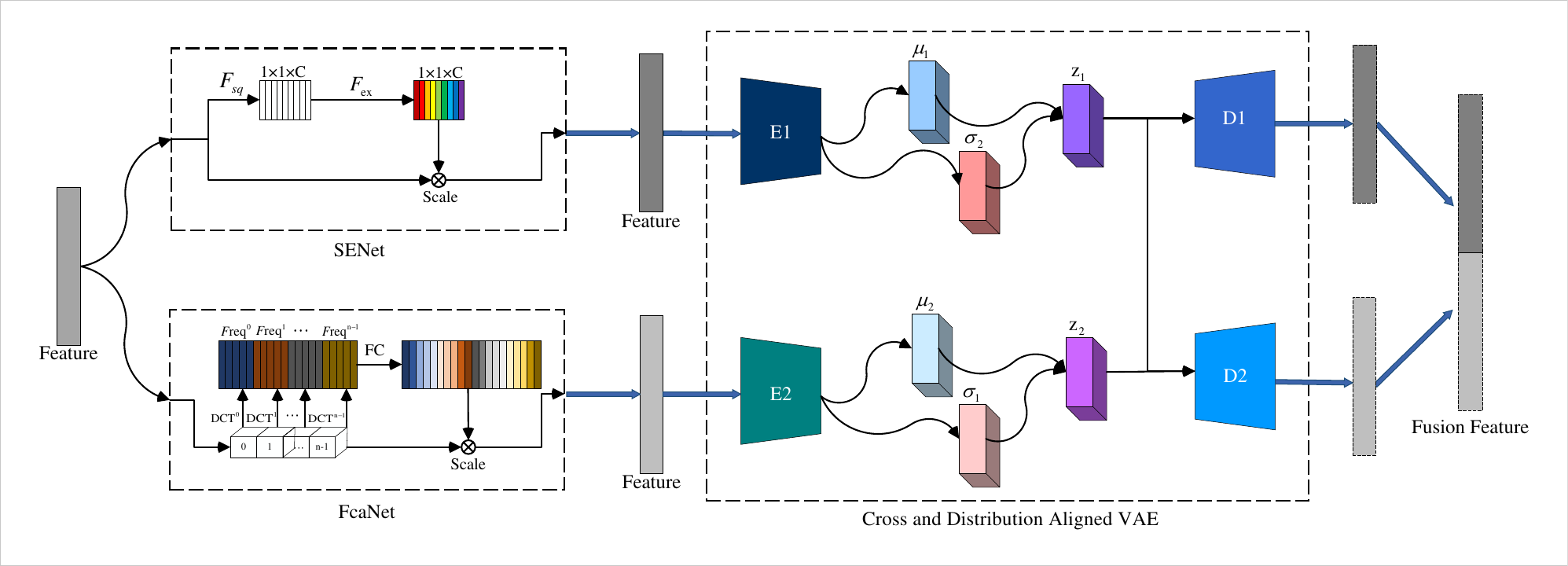}
		\end{center}
		\caption{\label{Figure 5}AFA module.}
	\end{figure}

	Spatial domain features refer to characteristics directly observed in the original pixel space of an image. These features, calculated based on the image's pixel values, include aspects such as color and brightness, edges and textures, as well as shape and geometric characteristics. In image processing and computer vision tasks, spatial domain features are crucial for understanding image content and conducting relevant analysis and feedback. The structural diagram of the spatial domain feature extraction (SFE) module is illustrated in Figure \ref{Figure 4}, which includes a ResNet12 backbone and a spatial-frequency domain attention feature alignment (AFA) module (illustrated in Figure \ref{Figure 5}).
	
	\subsubsection*{Backbone}
	
	Firstly, the 3${\times}$84${\times}$84 input image is fed into a CNN with ResNet12 as the backbone network. After passing through four residual blocks, the dimensions of the output feature are 640${\times}$5${\times}$5. Secondly, after average pooling, a 640${\times}$1${\times}$1 feature vector is generated. Finally, this one-dimensional vector is input into the AFA module to produce the final required spatial domain feature vector.
	
	\subsubsection*{Spatial-frequency domain attention feature alignment module: AFA}
	
	AFA incorporates the spatial attention mechanism SENet\cite{hu2018squeeze} and the frequency domain attention mechanism FcaNet\cite{qin2021fcanet}, along with the cross-alignment mechanism CADA-VAE\cite{schonfeld2019generalized}, this paper introduces a spatial-frequency domain attention feature alignment network. At first, the image features extracted by the convolutional neural network are sent to the spatial attention module SENet and the frequency domain attention module FcaNet separately. Combining SENet and FcaNet for the same input image allows effective utilization of spatial correlations and frequency domain distribution. Through selective weighting of spatial and frequency domain features, the network enhances its attention to distinct spatial and frequency domain features, which helps compensate for the limitations of individual attention mechanisms, guiding the network to extract more discriminative and comprehensive features. Secondly, the output features from both branches are simultaneously input into the alignment mechanism, where useful feature information within the channels is fused. This provides an enhancement of the model's generalization ability, and robustness of diverse image inputs, thereby improving the network's capacity to comprehend the features of input samples.
	
	\subsubsection*{Cross and Distribution Aligned VAE}
	
	The Cross and Distribution Aligned VAE (CADA-VAE) are applied to align spatial and frequency attention-weighted features of the same image through three main modules: cross alignment, distribution alignment, and Variational Auto-Encoder (VAE)\cite{kingma2013auto}. Among them, the loss function of VAE is:
	\begin{equation}
		\begin{aligned}
			{{\cal L}_{VAE}} = & \sum\limits_i^M {{E_{{q_\phi }(z|x)}}[\log {p_\theta }({x^{(i)}}|z)]} - \varepsilon {D_{KL}}({q_\phi }(z|{x^{(i)}})||{p_\theta }(z)),
		\end{aligned}
	\end{equation}
	where M represents input in different forms. The encoder is a neural network, the input is the feature x, the output is the hidden vector z, and the parameter is $\varphi $. Therefore, ${q_\varphi }(z|x)$ represents the encoding process from x to z. The decoder is also a neural network, the input is the hidden vector z, the output is the probability distribution of the feature, and the parameter is $\theta $. Therefore, ${p_\theta }(x|z)$ represents the decoding process from z to x. $p(z)$ represents the standard normal distribution, and $\varepsilon $ represents the weight of KL divergence. The CA loss function is:
	\begin{equation}
		\begin{aligned}
			{{\cal L}_{CA}} = \sum\limits_i^M {\sum\limits_{j \ne i}^M {|{x^{(j)}} - {D_j}({E_i}({x^{(i)}}))|} },
		\end{aligned}
	\end{equation}
	where ${D_j}$ is the j-th decoder and ${E_i}$ is the ith encoder. The DA loss function is:
	\begin{equation}
		\begin{aligned}
			{{\cal L}_{DA}} = \sum\limits_i^M {\sum\limits_{j \ne i}^M {{W_{ij}}} },
		\end{aligned}
	\end{equation}
	\begin{equation}
		\begin{aligned}
			{W_{ij}} = {(||{\mu _i} - {\mu _j}||_2^2 + ||\Sigma _i^{\frac{1}{2}} - \Sigma _j^{\frac{1}{2}}||_{Frobenius}^2)^{\frac{1}{2}}},
		\end{aligned}
	\end{equation}
	where ${W_{{\rm{ij}}}}$ represents the Wasserstein distance \cite{rubner2000earth} between the i-th encoder and the j-th decoder, which is mainly used to describe the difference between the two distributions. The encoder predicts $\mu $ and $\sigma $ such that ${q_\varphi }(z|x) = {\rm N}(\mu,\sigma )$ holds, thereby generating the intermediate vector z through the reparameterization technique. The total loss function consists of three parts: VAE loss, CA loss, and DA loss:
	\begin{equation}
		\begin{aligned}
			{{\cal L}_{CADA - VAE}} = {{\cal L}_{VAE}} + \alpha {{\cal L}_{CA}} + \beta {{\cal L}_{DA}},
		\end{aligned}
	\end{equation}
	where $\alpha $ and $\beta $ represent the weighting factors of cross-alignment loss and distribution alignment loss respectively.
	
	\subsection*{Frequency-Domain Feature Extraction Module: FFE}
	
\begin{figure}[ht]
		\begin{center}
			\includegraphics[width=1.0\linewidth]{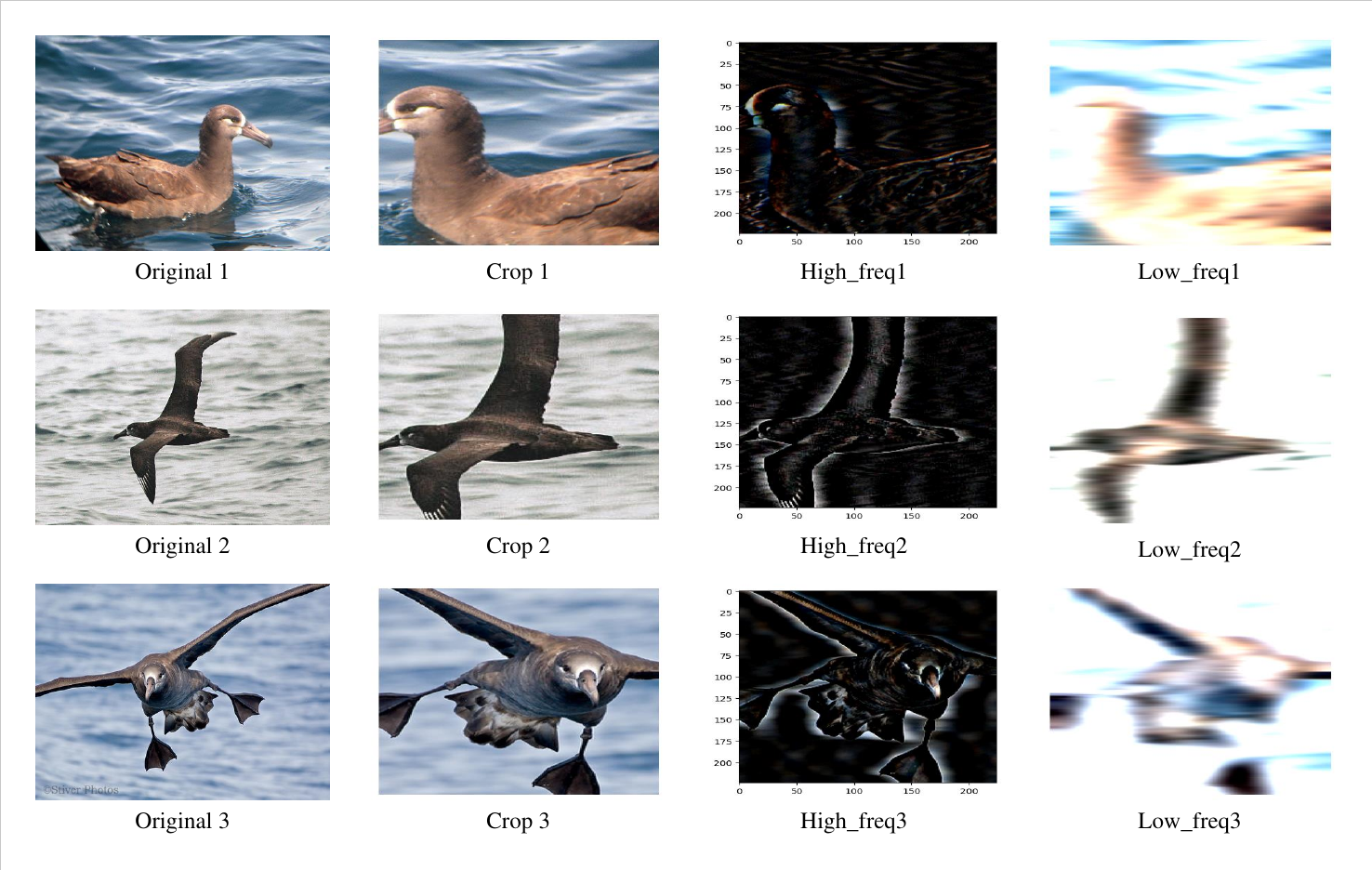}
		\end{center}
		\caption{\label{Figure 6}High and low frequency components of the sample images.}
	\end{figure}

	Research indicates that spatial features and frequency domain features of an image are two distinct ways of describing its content. They capture information from different perspectives. To express the rich feature information of images as much as possible, this paper proposes the Frequency Domain Feature Extraction (FFE) module to further extract frequency domain features. For a sample image, high-frequency information in the image's frequency domain features typically represents details and edge information, while low-frequency information represents overall brightness and color distribution in the image. As shown in Figure \ref{Figure 6}, it represents the high and low-frequency components of the sample image. Due to the distinctiveness of high and low-frequency components, we choose to assign different scale coefficients to the high and low-frequency components of the image. This further captures important information in the image, leading to the extraction of more comprehensive image features.
	
\begin{figure}[ht]
		\begin{center}
			\includegraphics[width=1.0\linewidth]{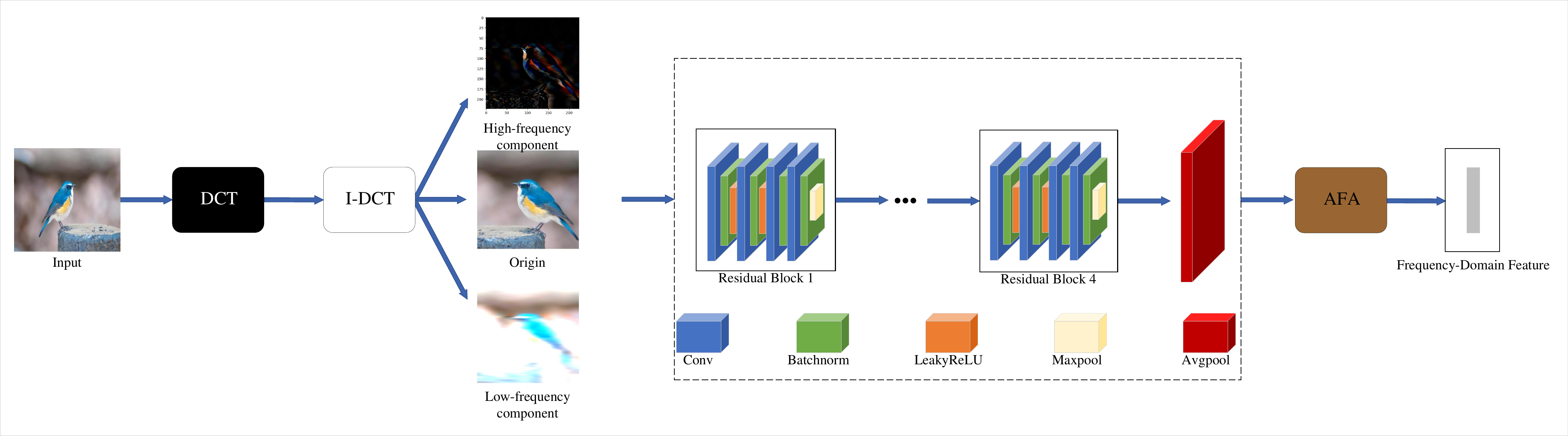}
		\end{center}
		\caption{\label{Figure 7}FFE module.}
	\end{figure}

	The frequency-domain feature extraction (FFE) module also uses ResNet12 as the backbone network. The network structure is shown in Figure \ref{Figure 7}. Firstly, a Discrete Cosine Transform (DCT) is applied to the input image to generate its frequency spectrum. Secondly, as the low-frequency part is located in the upper left corner of the array and the high-frequency part is located in the lower right corner of the array, a frequency coefficient selection is utilized to generate the high-frequency component spectrum and low-frequency component spectrum separately. Following that, the DCT inverse transform is applied to each of the three spectrum images, resulting in reconstructed images corresponding to the original, high-frequency, and low-frequency components. Then, the three obtained reconstructed images are collectively input into the ResNet12 to generate a one-dimensional feature vector. Finally, this one-dimensional vector is input into the AFA module based on spatial frequency domain attention to generate the required final frequency domain feature vector.
	
	Support I(m,n) is a pixel at position (m,n) on the a N$\times$N image I, the two-dimensional Discrete Cosine Transform (2D DCT) \cite{ahmed1974discrete} can transform the two-dimensional discrete image I into a set of coefficients of discrete cosine functions, thereby achieving the conversion from spatial domain to frequency domain.
	\begin{equation}
		\begin{aligned}
			F(k,l) = & C(k)C(l)\sum\limits_{m = 0}^{N-1} \sum\limits_{n = 0}^{N - 1} {f(m,n)} \cos (\frac{{(2m + 1)k\pi }}{{2N}}) \cos (\frac{{(2n + 1)l\pi }}{{2N}}), k=0,1,...N - 1,
		\end{aligned}
	\end{equation}
	\begin{equation}
		\begin{aligned}
			C(k) = C(l) = \left\{ {\begin{array}{*{20}{c}}
					{\sqrt {\frac{1}{N}}, k,l = 0}\\
					{\sqrt {\frac{2}{N},} \text{else}}
			\end{array}} \right.
		\end{aligned}
	\end{equation}
	where $F(k,l)$ represents the value of the kth, lth frequency component of the signal or image in the frequency domain, while $f(m,n)$ represents the value of the mth, nth sample point of the signal or image in the time domain. N denotes the length of the signal or image, and C(k) and C(l) are scaling coefficients used for normalization.
	By performing the inverse operation on the transformed frequency domain coefficients, data can be converted back from the frequency domain to the spatial domain.
	\begin{equation}
		\begin{aligned}
			{\rm{f}}(m,n)=&\sum\limits_{k=0}^{N-1} \sum\limits_{l=0}^{N-1}{C(k)C(l)F(k,l)}\cos (\frac{{(2m+1)k\pi}}{{2N}})\cos(\frac{{(2n+1)l\pi}}{{2N}}), {\rm{m}},n=0,1,...N-1.
		\end{aligned}
	\end{equation}
	
	\subsection*{Spatial-Frequency Domain Feature Fusion}
	
	After obtaining the spatial and frequency domain features from SFE and FFE for each new sample data, CADA-VAE is used to align these features. Output-aligned features are concatenated as the final feature of each sample image. This feature includes rich spatial frequency domain information, enabling a more comprehensive capture of image details, and thereby improving model performance.
	
	\subsection*{zero-shot translation}
	
\begin{figure}[ht]
		\begin{center}
			\includegraphics[width=1.0\linewidth]{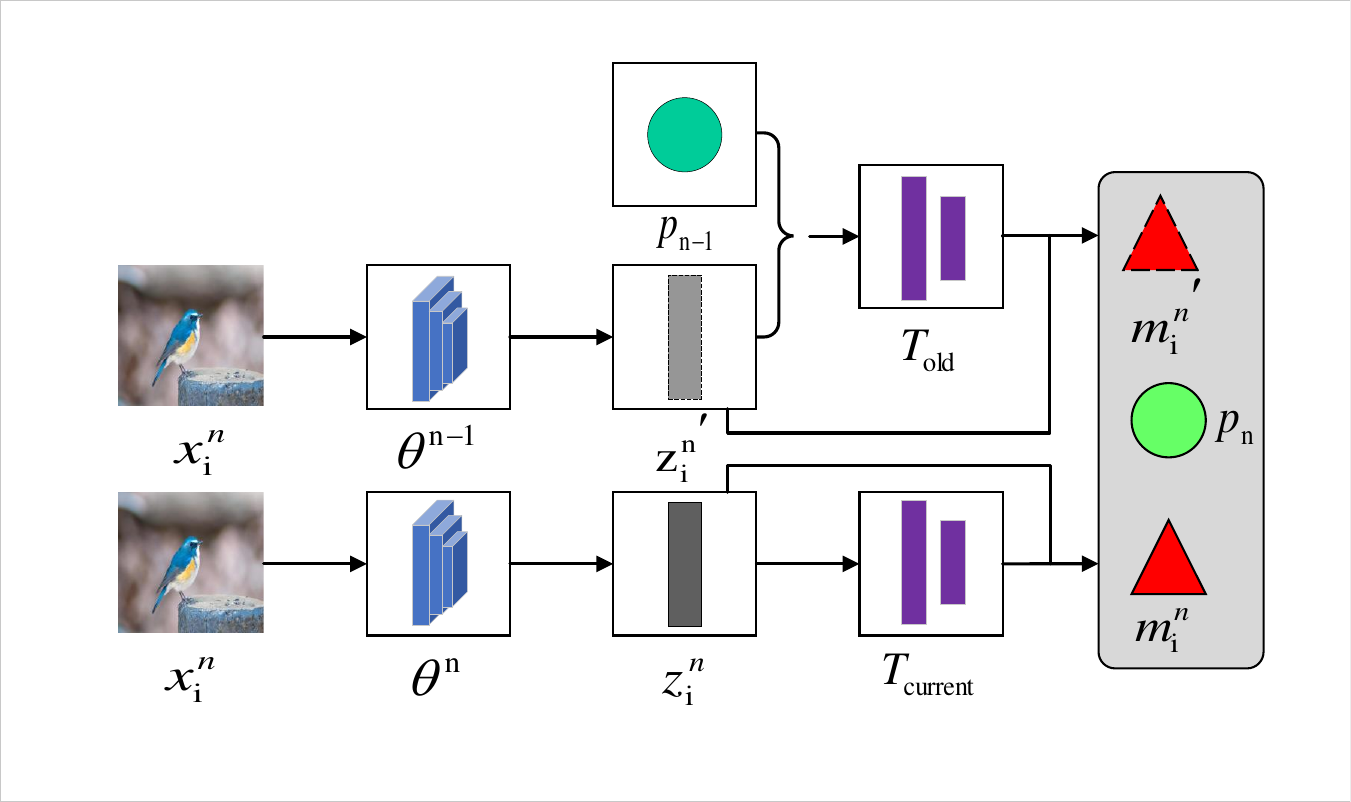}
		\end{center}
		\caption{\label{Figure 8}Zero-shot translation module.}
	\end{figure}

	For the n-th task, to exclude the use of samples from the (n-1)th task, we input the sample image $x_{\rm{i}}^n$ from the n-th task into the feature extraction models of both the (n-1)-th model ${\theta ^{{\rm{n}} - 1}}$ and current model ${\theta ^{\rm{n}}}$, obtaining features ${\rm{z}}{_{\rm{i}}^{\rm{n}}}^\prime$ and $z_{\rm{i}}^n$, respectively. Simultaneously, utilizing the prototype ${p_{{\rm{n}} - 1}}$ of the previous task, the zero-shot translation model ${T_{{\rm{old}}}}$ and ${T_{{\rm{current}}}}$ (shown in Figure \ref{Figure 8}) are employed to compute the compensated $m{_{\rm{i}}^n}^\prime $ and $m_{\rm{i}}^n$, and then update the prototype of the current task ${p_{\rm{n}}}$. Where $m{_{\rm{i}}^n}^\prime $ and $m_{\rm{i}}^n$ are defined as:
	\begin{equation}
		\begin{aligned}
			m{_{\rm{i}}^n}^\prime = {\rm{z}}{_{\rm{i}}^{\rm{n}}}^\prime + {T_{{\rm{old}}}}({\rm{z}}{_{\rm{i}}^{\rm{n}}}^\prime),
		\end{aligned}
	\end{equation}
	\begin{equation}
		\begin{aligned}
			m_{\rm{i}}^n = {\rm{z}}_{\rm{i}}^{\rm{n}} + {T_{{\rm{current}}}}({\rm{z}}_{\rm{i}}^{\rm{n}}).
		\end{aligned}
	\end{equation}
	
	After obtaining $m{_{\rm{i}}^n}^\prime $ and $m_{\rm{i}}^n$, we designed a compensation loss function ${{\cal L}_{compensation}}$ to compensate for prototype offset during the training process.
	\begin{equation}
		\begin{aligned}
			{{\cal L}_{compensation}} = \frac{1}{{{t^{\rm{n}}}}}\sum\limits_{i = 1}^{{t^n}} {||m{{_i^n}^\prime }}  - m_i^n|{|_1}.
		\end{aligned}
	\end{equation}
	
	After the training process of the zero-shot translation models, we employ ${T_{{\rm{old}}}}$ to update ${p_{{\rm{n}} - 1}}$ and ${T_{{\rm{current}}}}$ to update ${p_{\rm{n}}}$. Thus, the prototypes in prototype memory belong to the common embedding space.
	
	\subsection*{Loss Function}
	
	For different tasks, we iteratively update data, models, and zero-shot translation, refining prototypes for new and old classes. Regularization is applied to compensate for prototype offset. Therefore, the overall loss function for the entire network is defined as:
	\begin{equation}
		\begin{aligned}
			{{\cal L}_{ALL}} = {{\cal L}_{CADA - VAE}} + {{\cal L}_{compensation}}.
		\end{aligned}
	\end{equation}
	
	\section*{Experiment}
	
	In this section, we introduce the involved datasets, evaluation metrics, and implementation details. Subsequently, we compare the experimental results with the current state-of-the-art incremental learning methods to establish the effectiveness of our proposed method. Finally, we present ablation studies to demonstrate the effectiveness of different modules.
	
	\subsection*{Datasets}
	
	We use the standard datasets CUB-200-2011 \cite{wah2011caltech} and CIFAR-100 \cite{krizhevsky2009learning} for experiments. CUB-200-2011 is a collection of bird images, comprising 11,788 bird images distributed across 200 bird categories and 312 attributes. It is a fine-grained dataset. The CIFAR-100 dataset has 100 classes, with each class containing 600 color images sized 32$\times$32, making it a coarse-grained dataset.
	
	\subsection*{Implementation Details}
	
	All models are implemented in PyTorch and optimized using the Adam optimizer. ResNet-12 is employed as the pre-trained backbone network. The image size for CUB-200-2011 is adjusted to 256x256, while the image size for CIFAR-100 is adjusted to 32$\times$32 and then randomly cropped and flipped. In this setup, the number of epochs is set to 50, the batch size is set to 64, and the learning rates for CUB-200-2011 and CIFAR-100 are set to 1e-5 and 1e-6, respectively.
	
	\subsection*{Evaluation Metric}
	
	This paper selects average incremental accuracy \cite{aljundi2017expert} and average forgetting \cite{chaudhry2018riemannian} as the evaluation metrics of the experiment. Among them, the average incremental accuracy reflects the overall performance of the test set in the entire task. ${a_{{\rm{k}},j}} \in [0,1]$ is defined as the accuracy of the j-th task after training k tasks in sequence, where j$\leq$k, and the average incremental accuracy of the k-th task is expressed as ${A_{\rm{k}}} = \frac{1}{k}\sum\limits_{j = 1}^k {{a_{k,j}}} $. The average forgetting is to estimate how much learned knowledge the network has forgotten. The forgetting of the j-th task is defined as $f_{\rm{j}}^k = \mathop {\max }\limits_{l \in 1,...,k - 1} ({a_{l,j}} - {a_{k,j}}),\forall j < {\rm{k}}$, and the average forgetting of the k-th task is defined as ${F_{\rm{k}}} = \frac{1}{{k - 1}}\sum\limits_{j = 1}^{k - 1} {f_j^k} $.
	
	\subsection*{Baseline Methods}
	
	In recent years, researchers have proposed many methods to overcome catastrophic forgetting during incremental learning. we append an “E” to the name of the original method designed for a classification network, E-FT would be FT (Fine-Tuning) adapted for an embedding network.
	
	E-FT: In order to compare the classification space and embedding network set during the incremental learning process, finetuning is used to tune the new tasks. One is to fix the parameters of the output layer and use a low learning rate to learn the parameters of the new task. The other is to fix the parameters of the convolutional layer to prevent overfitting and optimize the parameters of the fully connected layer.
	
	E-LWF: In order to match the softmax output of the old task network on the new task when the same input is determined. They want to ensure that the output embedding of the current task (${\rm{z}}_{\rm{i}}^t$) is similar to the output embedding of the previous task (${\rm{z}}_{\rm{i}}^{t - 1}$), constraining the parameter update implementation by minimizing the distance between the output embeddings of the image. The loss function is as follows:
	\begin{equation}
		\begin{aligned}
			{{\cal L}_{LWF}} = ||z_i^t - z_i^{t - 1}||,
		\end{aligned}
	\end{equation}
	where, $| \cdot ||$ represents the Frobenius norm.
	
	E-ECA: When training a new task, the current parameters are close to the optimal parameters in the previous task and can be used in the embedding network. The objective function is as follows:
	\begin{equation}
		\begin{aligned}
			{{\cal L}_{EWC}} = \sum\limits_p {\frac{1}{2}} {\cal F}_p^{t - 1}{(\theta _p^t - \theta _p^{t - 1})^2},
		\end{aligned}
	\end{equation}
	where, ${{\cal F}^{{\rm{t - }}1}}$ is the Fisher information matrix calculated after learning the previous task (t-1) and summing all parameters ${\theta _p}$ of the network.
	
	E-MAS: Accumulate an importance measure for each parameter in the network based on the sensitivity of the predicted output function to parameter changes. The objective function is as follows:
	\begin{equation}
		\begin{aligned}
			{{\cal L}_{MAS}} = \sum\limits_p {\frac{1}{2}} {\Omega _{\rm{p}}}{(\theta _p^t - \theta _p^{t - 1})^2},
		\end{aligned}
	\end{equation}
	where ${\Omega _p}$ is the sensitivity to its change estimated by the square of the L2 norm of the output function.
	This loss can be added to the metric learning loss as a way to prevent forgetting as the embeddings are continuously trained. The overall loss function is as follows:
	\begin{equation}
		\begin{aligned}
			{\cal L} = {{\cal L}_{ML}} + \gamma {{\cal L}_C},
		\end{aligned}
	\end{equation}
	where, $C \in \{ LWF,EWC,MAS\} $, $\gamma $ is the balance coefficient between metric learning loss and other losses.
	
	SDC: In order to supplement several existing incremental learning methods, a semantic drift method that is similar to the prototype in the new task training process is proposed to further improve performance. The loss function is as follows:
	\begin{equation}
		\begin{aligned}
			{{\cal L}_T} = \max (0,{d_ + } - {d_ - } + m),
		\end{aligned}
	\end{equation}
	where ${d_ + }$ and ${d_ - }$ are the Euclidean distances between the embedding of anchor ${z_{\rm{a}}}$ and the positive instance ${z_{\rm{p}}}$ and negative instance ${z_{\rm{n}}}$ respectively.
	
	ZSTCI: Based on the zero-shot incremental learning method, a Zero-Shot Translation model is constructed with the purpose of obtaining a unified representation of all classes, which can capture and measure the distribution between classes. When combined with existing incremental learning methods, the performance of the model can be further improved.
	
	PFR: In order to solve the catastrophic forgetting problem and discriminatively learn features to recognize fine-grained images, a Pseudo-set Frequency Refinement (PFR) \cite{pan4495470pseudo} architecture is proposed. Frequency-based information is introduced to refine original features and improve image classification accuracy.
	
	\subsection*{Results and Analysis}
	
	Table \ref{tab1} and Table \ref{tab2} summarize the average incremental accuracy results of all compared methods and our method on CUB-200-2011 and CIFAR100 base datasets. It can be seen that the SFDNet-equipped method achieves relatively the best results in all tasks on both datasets. In addition, all the experimental results in the table prove the existence of catastrophic forgetting in the embedded network, and also prove that our proposed method is more conducive to mitigating the catastrophic forgetting of the network.
	\begin{table}[ht]
		\caption{The average incremental accuracy on CUB dataset}\label{tab1}
		\centering
		\begin{tabular}{|l|l|l|l|l|l|l|l|l|l|l|}
			\hline
			Method & T1	& T2 & T3 & T4 & T5 & T6 & T7 & T8 & T9 & T10 \\
			\hline
			E-FT & 88.9 & 75.3 & 68.4 & 60.2 & 57.4 & 49.8 & 45.0 & 39.6 & 39.4 & 36.9 \\
			E-LWF & 88.9 & 78.0 & 72.5 & 66.7 & 61.9 & 57.2 & 54.5 & 53.4 & 49.5 & 45.9 \\
			E-EWC & 88.9 & 76.9 & 68.3 & 62.2 & 60.0 & 57.0 & 55.5 & 51.0 & 53.3 & 51.1 \\
			E-MAS & 88.9 & 74.7 & 66.4 & 59.0 & 58.2 & 54.7 & 53.2 & 48.7 & 50.1 & 49.1 \\
			E-SDC & 88.9 & 79.8 & 71.5 & 67.0 & 63.5 & 60.7 & 59.5 & 58.7 & 56.5 & 55.7 \\
			E-ZSTCI & 88.9 & 80.1 & 72.7 & 68.9 & 65.8 & 61.9 & 61.2 & 59.9 & 59.0 & 58.1 \\
			PFR & 77.8 & 73.7 & 71.4 & 68.8 & 68.0 & 63.0 & 62.9 & 60.6 & 59.6 & 58.7 \\
			SFDNet & 78.4 & 73.3 & 73.4 & 71.0 & 68.4 & 68.1 & 64.5 & 63.2 & 63.3 & 59.9 \\
			E-SFDNet & 89.6 & 78.6 & 76.2 & 71.9 & 69.8 & 69.0 & 66.3 & 64.9 & 63.6 & 60.4 \\
			\hline
		\end{tabular}
	\end{table}
	\begin{table}[ht]
		\caption{The average incremental accuracy on CIFAR100 dataset}\label{tab2}
		\centering
		\begin{tabular}{|l|l|l|l|l|l|l|l|l|l|l|}
			\hline
			Method & T1	& T2 & T3 & T4 & T5 & T6 & T7 & T8 & T9 & T10 \\
			\hline
			E-FT & 91.2 & 72.4 & 65.0 & 50.4 & 46.1 & 14.8 & 12.2 & 10.2 & 8.4 & 6.6 \\
			E-LWF & 91.2 & 78.5 & 76.7 & 72.5 & 70.6 & 60.4 & 54.5 & 49.6 & 44.5 & 40.9 \\
			E-EWC & 91.2 & 78.5 & 76.1 & 73.0 & 71.1 & 59.6 & 50.6 & 34.7 & 15.3 & 10.3 \\
			E-MAS & 91.2 & 79.2 & 77.0 & 73.3 & 70.7 & 60.9 & 53.7 & 40.8 & 19.7 & 11.0 \\
			E-SDC & 91.2 & 78.6 & 76.7 & 72.7 & 70.7 & 61.0 & 55.5 & 50.7 & 45.4 & 42.0 \\
			E-ZSTCI & 91.2 & 78.8 & 77.1 & 73.1 & 71.1 & 62.3 & 57.4 & 53.2 & 49.0 & 46.1 \\
			PFR & 71.3 & 69.2 & 66.1 & 63.9 & 61.8 & 58.5 & 56.6 & 54.9 & 53.7 & 51.6 \\
			SFDNet & 73.0 & 70.4 & 67.4 & 65.4 & 63.5 & 60.1 & 58.5 & 56.7 & 55.6 & 53.7 \\
			E-SFDNet & 92.0 & 80.6 & 78.3 & 72.1 & 65.3 & 62.4 & 60.1 & 58.5 & 57.3 & 54.0 \\
			\hline
		\end{tabular}
	\end{table}
	On CUB-200-2011, the network equipped with SFDNet achieved an average incremental accuracy of 59.9${\%}$. Compared with the original baseline network E-FT, it has improved by 23.0${\%}$, and compared with the current state-of-the-art network, it has improved by 1.2${\%}$. On CIFAR100, the average incremental accuracy of the network equipped with SFDNet reached 53.7${\%}$. Compared with the original baseline network E-FT, it has improved by 47.1${\%}$, and compared with the current state-of-the-art network, it has improved by 2.1${\%}$. It can be seen from the data in the table that compared with the current state-of-the-art methods, the proposed SFDNet can better improve the semantic gap problem in two adjacent tasks, alleviate the catastrophic forgetting of the network, and thus effectively improve the model performance.
	
\begin{figure}[ht]
		\begin{center}
			\includegraphics[width=1.0\linewidth]{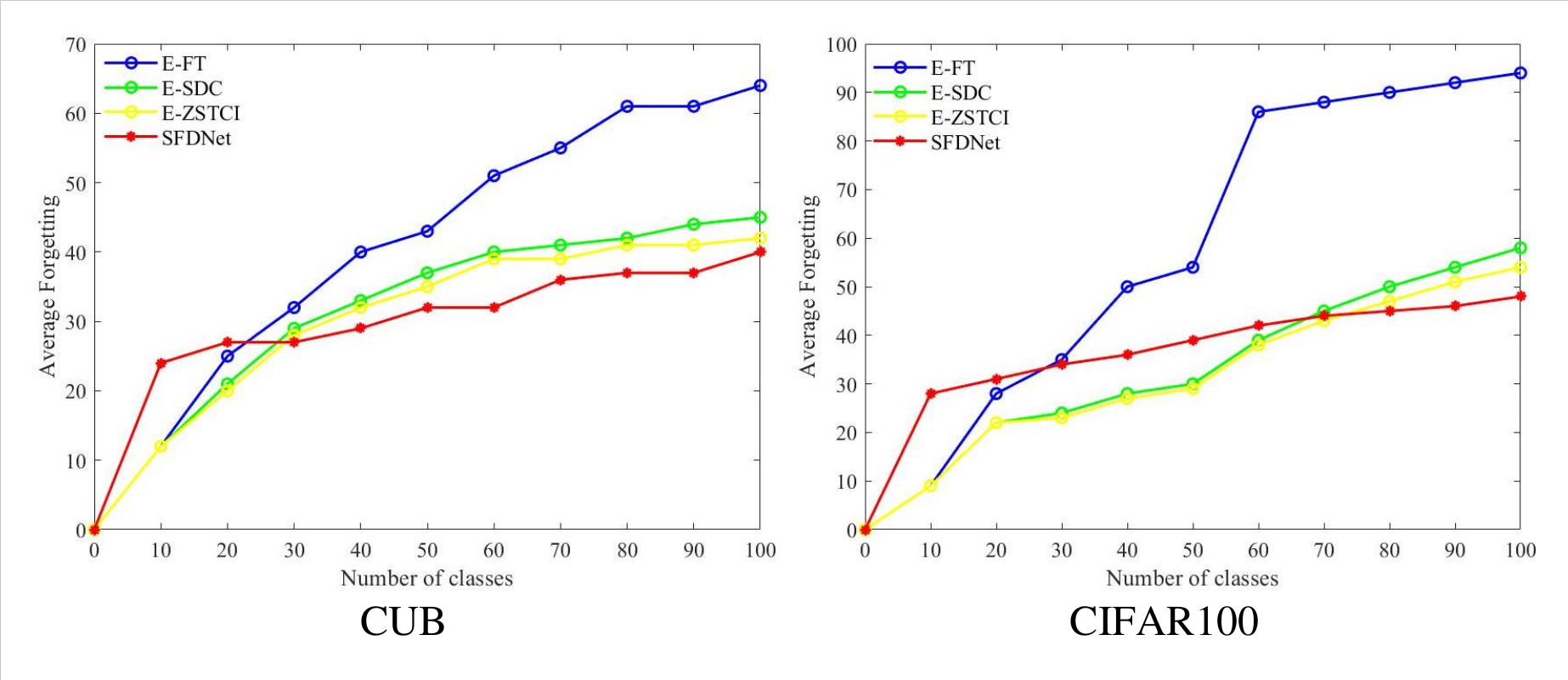}
		\end{center}
		\caption{\label{Figure 9}Average forgetting results}
	\end{figure}

	As shown in Figure \ref{Figure 9}, the average forgetting results on the two basic data sets CUB-200-2011 and CIFAR100 are shown. As can be seen from the figure, the average forgetting of all methods becomes evident as the number of categories increases, thus demonstrating the existence of catastrophic forgetting in the embedding network. However, after combining our proposed SFDNet method, the forgetting effect becomes less, once again proving that our proposed method can effectively alleviate catastrophic forgetting.
	
	\subsection*{Ablation Study}
	
	We conducted ablation experiments to study the effectiveness of our method, that is, the results of adding different modules to the basic model are shown in Table \ref{tab3}. The spatial-domain attention mechanism SENet, the frequency-domain attention mechanism FcaNet, the spatial-domain feature extraction (SFE) module and the frequency-domain feature extraction (FFE) module are separately added. According to the results in the table, when a certain module is added alone, although the effect is improved, it is not optimal. When SFDNet is added to the model, the effect is significantly improved, and the catastrophic forgetting of the network is also greatly alleviated.
	\begin{table}[ht]
		\caption{Ablation Study}\label{tab3}
		\centering
		\begin{tabular}{|l|l|l|}
			\hline
			Dataset & CUB & CIFAR100 \\
			\hline
			Base & 36.9 & 6.6 \\
			+SE & 51.2 & 41.1 \\
			+Fca & 50.6 & 41.5 \\
			+SFE & 58.3 & 49.1 \\
			+FFE & 58.7 & 51.6 \\
			+SFDNet(ours) & 59.9 & 53.7 \\
			\hline
		\end{tabular}
	\end{table}
	
	When only "SENet" is added to the basic model, compared with the baseline network, the CUB data set can be improved by up to 14.3${\%}$, and the CIFAR100 data set can be improved by up to 34.5${\%}$. When only "FcaNet" is added to the basic model, compared with the baseline network, the CUB data set can be improved by up to 13.7${\%}$, and the CIFAR100 data set can be improved by up to 34.9${\%}$. When only "SFE" is added to the basic model, compared with the baseline network, the CUB data set can be improved by up to 21.4${\%}$, and the CIFAR100 data set can be improved by up to 42.5${\%}$. When only "FFE" is added to the basic model, compared with the baseline network, the CUB data set can be improved by up to 21.8${\%}$, and the CIFAR100 data set can be improved by up to 45.0${\%}$. As can be seen from the data in the table, our proposed SFDNet can maximize the average incremental accuracy and alleviate the catastrophic forgetting of the network.
	
	\section*{Conclusion}
	
	This article proposes a zero-shot incremental learning algorithm based on spatial-frequency domain attention feature alignment to alleviate catastrophic forgetting in embedded networks. To reduce the semantic gap between new and old sample classes, a spatial-frequency domain attention feature alignment (AFA) module is introduced to capture salient information from different domains of each sample using spatial and frequency domain attention mechanisms. Additionally, a dual-domain feature extraction network (SFFE) is proposed to obtain more comprehensive feature information. Convolutional neural networks and discrete cosine transform are used to extract spatial and frequency domain features of images, followed by feature fusion. The fused features are then fed into the NCM classifier for image classification, fundamentally improving the effectiveness of feature extraction and reducing the distance between new and old tasks. Furthermore, our proposed method can be flexibly combined with other regularization-based incremental learning methods to further improve model performance. Experimental results demonstrate that our method outperforms state-of-the-art methods on two benchmark datasets.
	\bibliography{cite}
	
	%
	%
	%
	%
	%
	%
	%
	%
	
\end{document}